\documentclass[lettersize,journal]{IEEEtran}
\usepackage{amsmath,amsfonts}
\usepackage{array}
\usepackage[caption=false,font=normalsize,labelfont=sf,textfont=sf]{subfig}
\usepackage{textcomp}
\usepackage{stfloats}
\usepackage{url}
\usepackage{verbatim}
\usepackage{graphicx}
\def\BibTeX{{\rm B\kern-.05em{\sc i\kern-.025em b}\kern-.08em
    T\kern-.1667em\lower.7ex\hbox{E}\kern-.125emX}}
\usepackage{balance}

\usepackage{xspace}
\usepackage{booktabs}
\usepackage[numbers,sort&compress]{natbib}
\newcommand{\eg}{{\emph{e.g.}}}

\usepackage[normalem]{ulem} 
\useunder{\uline}{\ul}{}

\usepackage{algorithm}
\usepackage{algpseudocode}

\newcommand{\name}{BiXFormer}

\usepackage[switch]{lineno}

\usepackage{amssymb}

\usepackage{mathbbol}
\usepackage{multirow}
\usepackage{graphicx}
\usepackage{amssymb}
\usepackage{wrapfig}
\usepackage{caption}
\usepackage{subcaption}
\usepackage{multirow}

\usepackage{microtype}      
\usepackage{xcolor}         

\begin{document}
\title{BiXFormer: A Robust Framework for Maximizing Modality Effectiveness in Multi-Modal Semantic Segmentation}

\author{Anonomous Authors,
\thanks{Manuscript created October, 2020; This work was developed by the IEEE Publication Technology Department. This work is distributed under the \LaTeX \ Project Public License (LPPL) ( http://www.latex-project.org/ ) version 1.3. A copy of the LPPL, version 1.3, is included in the base \LaTeX \ documentation of all distributions of \LaTeX \ released 2003/12/01 or later. The opinions expressed here are entirely that of the author. No warranty is expressed or implied. User assumes all risk.}
}

\author{
Jialei Chen$^*$, Xu Zheng$^\dagger$, Danda Pani Paudel, Luc Van Gool, Hiroshi Murase,~\IEEEmembership{Life Fellow,~IEEE} Daisuke Deguchi,~\IEEEmembership{Member,~IEEE,} 
\thanks{Jialei Chen, Daisuke Deguchi, Hiroshi Murase are with the Graduate School 
of Informatics, Nagoya University, Nagoya, 
Japan. Xu Zheng is with AI Thrust, The Hong Kong University of Science and Technology, Guangzhou Campus (HKUST-GZ), Guangzhou, 
China. 
Danda Pani Paudel, Luc Van Gool are with the INSAIT, Sofia University, St. Kliment Ohridski.
* indicates the corresponding author. $\dagger$ indicates the project leader.}
\thanks{Manuscript created October, 2020; This work was developed by the IEEE Publication Technology Department. This work is distributed under the \LaTeX \ Project Public License (LPPL) ( http://www.latex-project.org/ ) version 1.3. A copy of the LPPL, version 1.3, is included in the base \LaTeX \ documentation of all distributions of \LaTeX \ released 2003/12/01 or later. The opinions expressed here are entirely that of the author. No warranty is expressed or implied. User assumes all risk.}
}

\markboth{Journal of \LaTeX\ Class Files,~Vol.~18, No.~9, September~2020}%
{How to Use the IEEEtran \LaTeX \ Templates}

\maketitle

\begin{abstract}
Utilizing multi-modal data enhances scene understanding by providing complementary semantic and geometric information. Existing methods fuse features or distill knowledge from multiple modalities into a unified representation, improving robustness but \textbf{restricting} each modality’s ability to fully leverage its strengths in different situations. We reformulate multi-modal semantic segmentation as a mask-level classification task and propose \textbf{BiXFormer}, which integrates Unified Modality Matching (UMM) and Cross Modality Alignment (CMA) to maximize modality effectiveness and handle missing modalities. Specifically, BiXFormer first categorizes multi-modal inputs into RGB and X, where ``X” represents any non-RGB modalities, \eg, depth, allowing separate processing for each. This design leverages the well-established pretraining for RGB, while addressing the relative lack of attention to X modalities. Then, we propose UMM, which includes Modality Agnostic Matching (MAM) and Complementary Matching (CM). MAM assigns labels to features from all modalities without considering modality differences, leveraging each modality's strengths. CM then reassigns unmatched labels to remaining unassigned features within their respective modalities, ensuring that each available modality contributes to the final prediction and mitigating the impact of missing modalities. Moreover, to further facilitate UMM, we introduce CMA, which enhances the weaker queries assigned in CM by aligning them with optimally matched queries from MAM. Experiments on both synthetic and real-world multi-modal benchmarks demonstrate the effectiveness of our method, achieving significant improvements in mIoU of \textbf{+2.75\%} and \textbf{+22.74\%} over the prior arts.

\end{abstract}


\section{Introduction}
\label{sec:intro}

\begin{figure}[t]
\begin{center}
\includegraphics[width=1\linewidth]{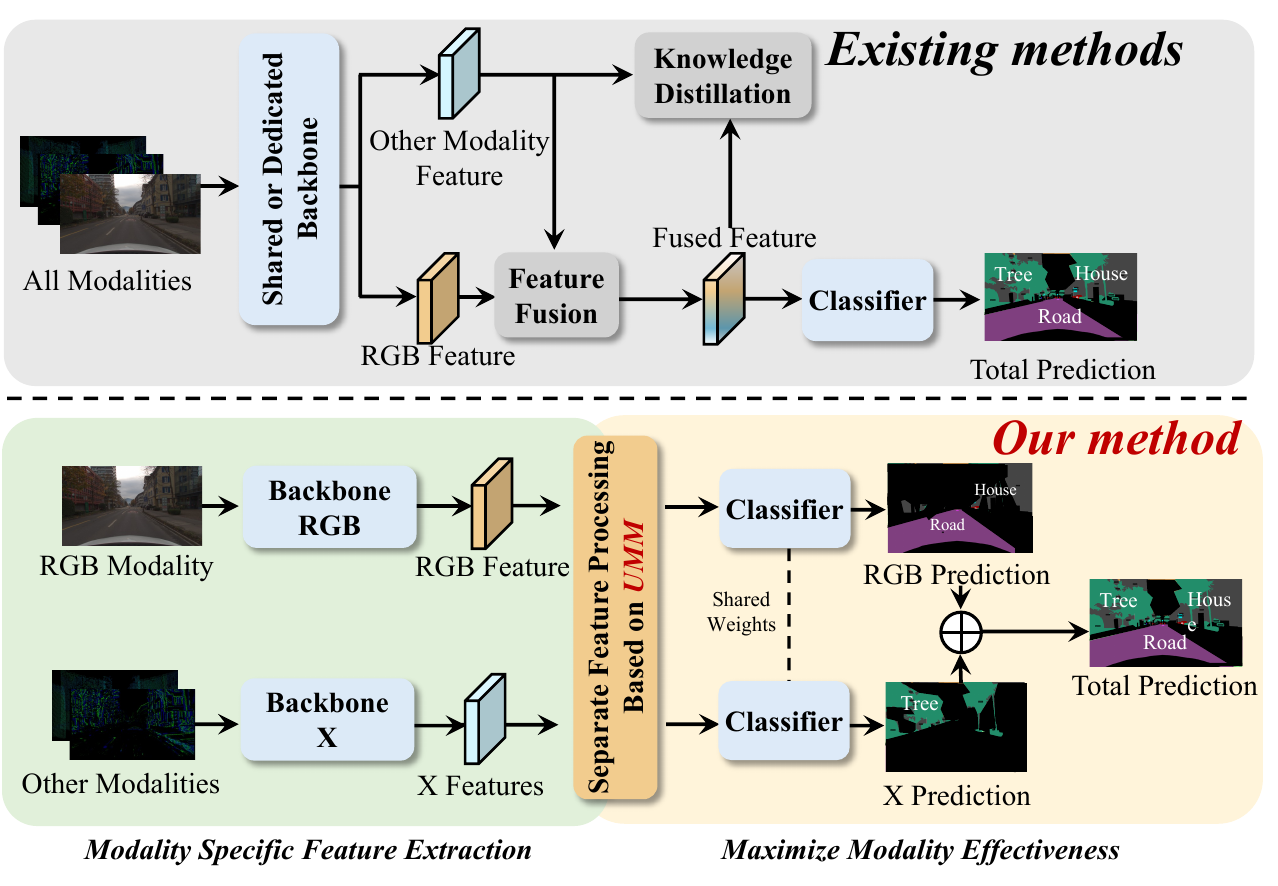}
\end{center}
\vspace{-0.2in}
\caption{Existing methods (top) rely on feature fusion and knowledge distillation, which limits the contribution of individual modalities. In contrast, our method (bottom) maximizes each modality’s (both RGB and non-RGB) contribution by using separate backbones and a unified modality matching mechanism. This allows modalities to focus on the labels they excels at.}
\label{fig:teaser}
\vspace{-0.3in}
\end{figure}

Semantic segmentation, a fundamental task in computer vision, aims to assign a class label to each pixel in an image \cite{FCN,deeplab,deeplabev3,setr,tcsvtseg1,zhong2025omnisam}. However, most existing methods rely solely on RGB images, which may be insufficient in real-world scenarios, such as high-speed motion or low-light conditions, where RGB information alone is often ambiguous or unreliable \cite{muses,deliver,nyuv2}. To address such limitation, increasing attention has been given to multimodal semantic segmentation \cite{magic,anyseg,multimodalityseg1,any2seg,deliver,cmx,zhao2025unveiling,liao2025benchmarking}, which incorporates complementary information from multiple modalities, such as depth, thermal, or event-based data, to enhance segmentation accuracy and robustness, leading to more reliable scene understanding across diverse environments.

Existing multimodal methods often rely on feature fusion \cite{deliver,xpromt,robustmultilearning,dformer,cpal,contextaware,embracingevent,feature-distillation3,tcsvtfusion2,tcsvtfusion3,tcsvtfusion4,tcsvtmulti1,zheng2025reducing} and knowledge distillation \cite{primkd,any2seg,anyseg,magic}. 
These two approaches can be combined to further improve performance by leveraging both feature fusion and knowledge transfer \cite{magic,any2seg} as shown in Fig. \ref{fig:teaser}.
However, these methods face two main limitations, including \textcircled{\textbf{1}} fusion-based methods adopt tightly coupled fused features and \textbf{\textit{fail to fully exploit the unique strengths of RGB and non-RGB modality}} for semantic segmentation, leading to suboptimal performance; and \textcircled{\textbf{2}} knowledge distillation methods rely heavily on the teacher model, but selecting an optimal teacher is challenging, as \textit{\textbf{fixed teacher models are often not well-suited}} to guide multimodal learning.
Therefore, a new multimodal framework is needed to fully leverage the strengths of RGB and non-RGB modality, enabling a more robust perception. Additionally, this framework must handle missing modality scenarios, which are common in real-world applications, such as the absence of depth sensors in low-cost systems due to hardware constraints, the unavailability of LiDAR in certain environments like indoor settings, or sensor failures under adverse weather conditions.

To \textbf{\textit{maximize RGB and non-RGB modalities' contribution}} and \textbf{\textit{mitigate the impact of missing modalities}}, we introduce \name, the \textbf{\textit{first}} robust framework designed in a loosely coupled manner, where RGB and non-RGB modality is processed independently, and their predictions rather than features are combined. This design enables RGB and non-RGB modalities to specialize in the labels they perform best at, leveraging the extensive pretraining available for RGB while compensating for the limited attention given to X modalities. First, to preserve RGB features when integrating non-RGB modalities, we employ two separate backbones: a backbone RGB pre-trained on ImageNet~\cite{imagenet} and a backbone X with a modified first layer to handle non-RGB modalities. Unlike shared-backbone methods, our approach prevents feature fusion by maintaining distinct backbones for RGB and non-RGB modality. Additionally, our design enables flexible multimodal learning with only two backbones, avoiding dedicating backbones for RGB and non-RGB modality.Second, we introduce Unified Modality Matching (UMM) to ensure effective label matching across modalities. 
Drawing inspiration from ~\cite{mask2former,maskformer}, we introduce UMM to assign the most suitable labels to RGB and non-RGB modality. 
Concretely, UMM assigns labels through a two-step process:
\textcircled{\textbf{1}} \textbf{Modality Agnostic Matching} (MAM) first assigns labels to the most relevant queries without modality constraints, ensuring both RGB and non-RGB modality learns from the categories it is best suited for. However, MAM alone may lead to a dominant modality acquiring most labels, resulting in performance degradation when certain modalities are missing.
\textcircled{\textbf{2}} To address this, we introduce \textbf{Complementary Matching} (CM), which reassigns unmatched queries to any remaining unassigned labels within RGB and non-RGB modality, ensuring that missing modality scenarios are effectively handled. Since CM operates on leftover queries, its matching is inherently less optimal than those in MAM. To refine these weaker queries, we introduce Cross Modality Alignment (CMA), which aligns weaker queries assigned in CM with the queries matched in MAM. 

Overall, BiXFormer eliminates feature fusion by directly combining segmentation results from different modalities at the prediction stage, in contrast to standard fusion-based models \cite{cmx,deliver} and knowledge-distillation approaches~\cite{magic,any2seg}.  Compared to methods that adopt one-to-many matching \cite{groupdetr,detrswithcollaborative,detrswithhybridmatching,msdetr}, modify the matching metrics \cite{stablematching,rankdetr}, or refine queries \cite{dinodetection,emo2,hybridproposal,saliencedetr} in one-step manner, our approach employs a two-step matching process to maximize the contribution of both RGB and non-RGB modality.
Extensive experiments demonstrate that BiXFormer achieves SOTA performance. To recap, our contributions are:

\noindent \textbf{(I)} We propose \textbf{\name}, a novel robust framework that introduces UMM, maximizing the contributions of both RGB and non-RGB modality through MAM while mitigating the impact of missing modalities via CM.

\noindent \textbf{(II)} We introduce CMA, which leverages the optimally matched queries from MAM to strengthen the weaker queries in CM, improving segmentation robustness.

\noindent \textbf{(III)} We evaluate \name~across multiple benchmarks, surpassing existing multimodal segmentation methods in both full-modality and missing-modality scenarios.


\section{Related Works}
\label{sec:related works}

\begin{figure*}[t]
\begin{center}
\includegraphics[width=0.99\linewidth]{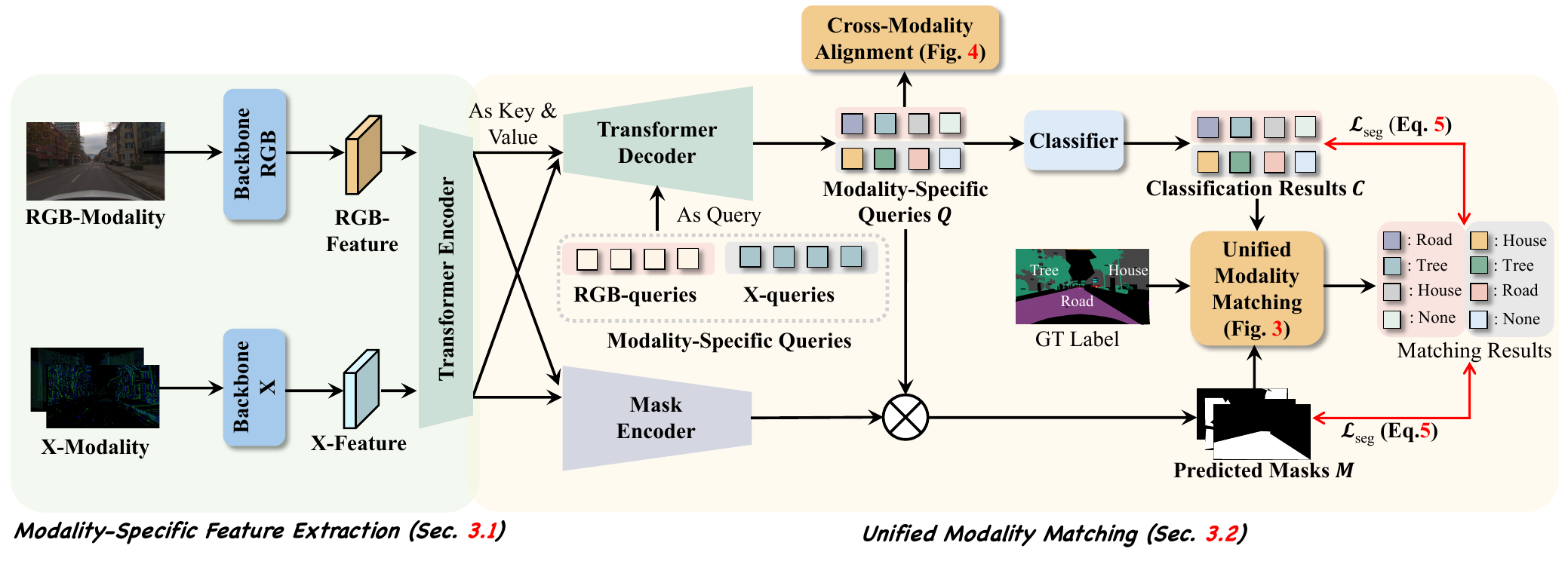}
\end{center}
\vspace{-0.25in}
\caption{Overview of our method. Given RGB images and images from X modalities, modality-specific features are extracted using their respective backbones: Backbone RGB and Backbone X. These features are then processed by a shared transformer encoder. Modality-specific queries are used as queries, while both RGB and X features serve as keys and values in the Transformer Decoder. The decoded features interact with the Mask Encoder before undergoing Unified Modality Matching (UMM), which assigns labels to modality-specific queries. To further refine the representations, Cross Modality Alignment (CMA) facilitates information transfer between modalities.}
\label{fig:main}
\vspace{-0.2in}
\end{figure*}

\subsection{Semantic Segmentation}

Semantic segmentation, as one of the most important tasks in computer vision, differs from image classification \cite{resnet,vggnet,convnext,zhu2024customize} in that it assigns a class label to each pixel of an image rather than one label to the entire image \cite{xie2023adversarial,zheng2023look,FCN,deeplab,deeplabev3,deformableconv,segformer,setr,segnext,frozen,zheng2022uncertainty,zheng2023both,zhu2023good,zheng2024semantics}. FCN \cite{FCN} was the first to treat segmentation as per-pixel classification using CNNs, providing a new perspective for subsequent works. However, FCN suffered from limited receptive field size. To address these issues, some works propose dilated convolution \cite{deeplab,deeplabev3}, and deformable convolution \cite{deformableconv}. With the rapid development of self-attention \cite{transformer,vit}, the receptive field is much enhanced due to the long-range dependency of the transformer. Besides the pixel-level classification view, some methods treat the semantic segmentation as a mask-level classification task \cite{mask2former,maskformer} and achieve impressive performance. Though successful, existing methods only rely on RGB images, which may fall short in some extreme environments, \eg, high-speed movement and low-light conditions. To our knowledge, this is the first attempt to marry the mask-classification segmentation framework with multi-sensor input. Previous multimodal segmentation models predict pixel-wise labels, whereas BiXFormer produces a set of mask proposals from both RGB and non-RGB modality and then assigns class labels to them in a coordinated fashion.

\subsection{Multi-modality Learning}
In real-world applications, relying on a single modality is often ineffective, particularly under extreme conditions. Consequently, multimodal learning has gained increasing attention, with numerous research efforts dedicated to this field \cite{zhou2023clip,lyu2024omnibind,zhou2024exact,magic,cmx,deliver,muses,xpromt,robustmultilearning,dformer,contextaware,embracingevent,primkd,missingmodalitrobustness,actionmae,unifiedmultimodalimagesynthesis,m3ae,tcsvtfusion2,tcsvtfusion,tcsvtfusion3,tcsvtfusion4,tcsvtmulti1,feature-distillation3,zheng2023deep,cao2024chasing,lyu2024unibind}. Existing approaches can be broadly categorized into two paradigms: fusion-based methods and knowledge distillation-based methods. Fusion-based methods extract features from RGB and other modalities using either shared \cite{magic,deliver,cmx,xpromt,zheng2024eventdance,zhou2024eventbind} or modality-specific \cite{embracingevent,contextaware,liao2025memorysam} backbones. They then integrate these features through various techniques, such as self-attention mechanisms \cite{xpromt,cmx}, to construct a robust multimodal representation. On the other hand, knowledge distillation-based methods focus on transferring beneficial knowledge to enhance feature representations \cite{magic,any2seg,zheng2024magic++}. Despite their effectiveness, these approaches face notable limitations. Fusion-based methods may lead to suboptimal feature representations by failing to fully exploit different modalities. Meanwhile, knowledge distillation-based methods often struggle with selecting an appropriate teacher model, and their representations may not fully capture the richness of each modality. Consequently, we propose BiXFormer, a novel framework that maximizes the contribution of RGB and non-RGB modality without relying on any pretrained teacher models.

\subsection{Matching Strategies for Query-Based Models}
To address the challenge of redundant predictions in object detection, traditional methods rely on Non-Maximum Suppression (NMS) \cite{FCOS,focalloss} to filter overlapping results. DETR \cite{detr} introduces a set-based matching mechanism with discrete object queries, enabling the model to directly identify potential objects in an image without the need for post-processing. These queries are assigned labels via Hungarian matching, which optimizes matchings based on matching losses. This paradigm has since been extended to segmentation tasks, enabling universal segmentation \cite{mask2former,maskformer,maskdino}. However, conventional Hungarian matching often suffers from suboptimal suitability, limiting overall performance. To address this, some methods introduce one-to-many matching \cite{groupdetr,detrswithcollaborative,detrswithhybridmatching,msdetr}, where a single label is assigned to multiple queries, leveraging the advantages of traditional approaches. Others refine the matching process by modifying matching metrics \cite{stablematching,rankdetr} to ensure more stable matchings or by enhancing query representations \cite{dinodetection,emo2,hybridproposal,saliencedetr} to improve label matching quality. 

In this paper, we propose Unified Modality Matching (UMM), a novel approach that incorporates two sequential matching steps to maximize the contribution of RGB and non-RGB modality while preserving the principles of conventional matching. Unlike prior works that primarily focus on refining matching stability or query representation within a single modality, UMM explicitly addresses modality interactions by first performing matching across modalities and then refining matching within RGB and non-RGB modality, ensuring a balanced contribution and mitigating performance degradation in missing modality scenarios.


\section{Methods}
\label{sec:methods}


\noindent \textbf{Preliminary.} Our framework processes multimodal visual data. Given a dataset $\mathbb{D} = \big\{\{\textbf{I}_{ij},\textbf{Y}_j\}_{i = 0}^{O}\big\}_{j=0}^{N}$, where $\textbf{I}_{ij} \in \mathbb{R}^{C_i\times H \times W}$ represents the data from the $i$th modality of the $j$th sample, $\textbf{Y}_j$ is the corresponding pixel-level annotation, $H$ and $W$ denote the height and width of the image, respectively. $N$ indicates the dataset size, $C_i$ indicates the channel number of the $i$th modality, and $O$ represents the number of modalities. The ground truth set is defined as $\textbf{Y} = \big\{ \textbf{y}^k = (c^k_y, \textbf{m}_y^k) \big\}_{k=0}^{U}$
where \( c^k_y \in [0, K] \) represents the class label, $U$ indicates the number of unique classes in an image, and \( \textbf{m}_y^k \in [0,1]^{H \times W} \) denotes the corresponding mask. To improve clarity, we define two functions: $C(\textbf{y})$ extracts the class label $c$ from the annotation $\textbf{y}$, and $M(\textbf{y})$ extracts the class-agnostic mask associated with $\textbf{y}$. For simplicity, we define the first modality as the RGB modality. We follow the data processing pipeline of~\cite{deliver}, where the channel of all modalities is set as 3. Besides, the field of view has been aligned for each modality. During training, each batch consists of samples from all modalities.

\noindent \textbf{Method Overview.} Our method maximizes the contribution of RGB and non-RGB modality by processing them separately and assigning suitable labels accordingly. The overall framework is illustrated in Fig.~\ref{fig:main}. First, we employ two backbones to extract features from RGB and non-RGB (X) modalities. Then, to leverage their distinct characteristics, we introduce \textbf{Unified Modality Matching (UMM)} for optimal label matching. Finally, to further enhance UMM, we introduce \textbf{Cross Modality Alignment (CMA)}, which aligns the best-matching queries from MAM with their suboptimal counterparts in Complementary Matching (CM), enabling information transfer between modalities to improve query representations.


\subsection{Modality Specific Feature Extraction}

Existing methods can be broadly categorized into shared-backbone approaches \cite{any2seg,magic,xpromt} and modality-specific-backbone approaches \cite{actionmae,unifiedmultimodalimagesynthesis}. Shared-backbone methods use a single ImageNet-pretrained model for all modalities and sometimes incorporate vision-language alignment for RGB. However, they are biased toward RGB features and often underperform on non-RGB inputs due to the lack of dedicated pre-training. Modality-specific methods assign separate backbones to RGB and non-RGB modalities, improving feature extraction but increasing training cost and complexity as the number of modalities grows. To balance these trade-offs, we adopt a dual-backbone design: one backbone for RGB and another shared across all non-RGB (X) modalities. This design leverages the mature pretraining paradigm for RGB \eg, ImageNet), while compensating for the limited pretraining resources of non-RGB modalities, thus supporting more balanced learning.

To maximize the contribution of RGB and non-RGB modality, we adopt a modality-specific feature extraction strategy. Unlike existing methods \cite{any2seg,magic,cmx,deliver} that employ a single shared backbone across all modalities, we explicitly separate them into RGB and non-RGB (X) modalities. Specifically, the RGB image is denoted as \( \textbf{I}_{r} = \textbf{I}_{0} \in \mathbb{R}^{3 \times H \times W} \), while the X modalities are represented as a set of images \( \{\textbf{I}_{i}\}_{i=1}^{O} \). To ensure a comprehensive representation of the X modalities, we concatenate them along the channel dimension, forming a unified input $\textbf{I}_{x} \in \mathbb{R}^{\big((O-1) \times 3\big) \times H \times W}$.
Instead of using a dedicated backbone for each modality which would lead to a linear increase in parameters with the number of sensors, we use a shared backbone X, leveraging the assumption that generic spatial features can be extracted.

To be spcific, we employ two independent backbones: Backbone RGB and Backbone X. The RGB backbone is initialized with an ImageNet-pretrained \cite{imagenet} model and processes \( \textbf{I}_r \) directly. For Backbone X, we also start with an ImageNet-pretrained model but \textit{modify its first convolutional layer to accommodate the concatenated X modalities}. Specifically, we replace the original first layer (designed for standard RGB of shape \( 3 \times C_{\text{out}} \), where \( C_{\text{out}} \) is the number of output channels) with a randomly initialized layer of shape: $\big( (O - 1) \times 3\big) \times C_{\text{out}}$.

\begin{figure}[t]
\begin{center}
\includegraphics[width=1\linewidth]{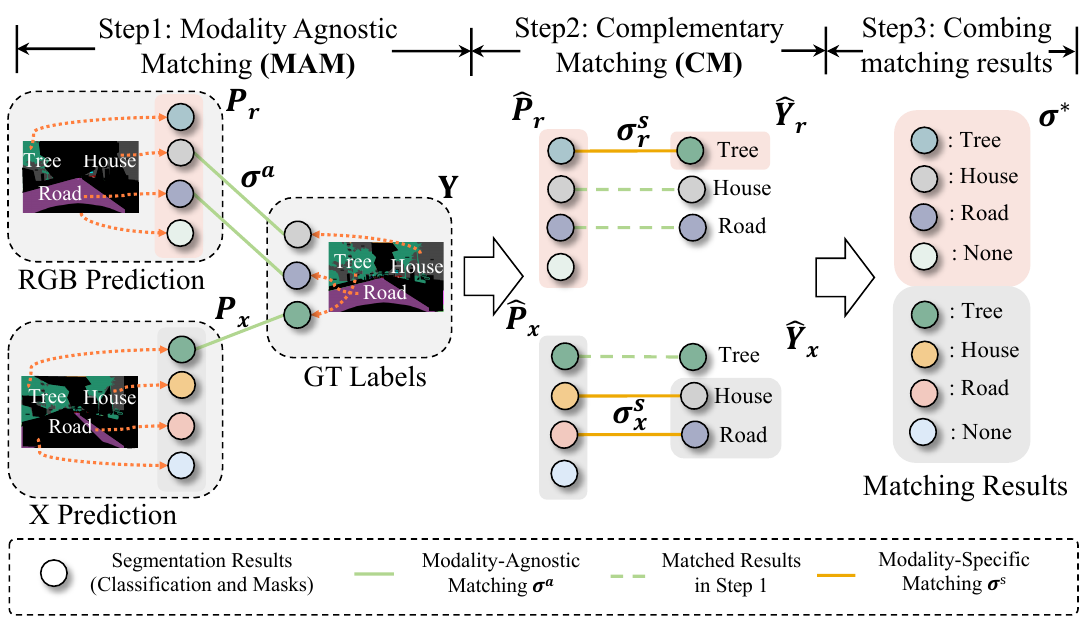}
\end{center}
\vspace{-0.1in}
\caption{Overview of the Unified Modality Matching (UMM), where the red dashed line indicates the correspondence between predictions and their respective areas.}
\label{fig:umm}
\vspace{-0.2in}
\end{figure}

Once initialized, Backbone RGB and Backbone X extract dense modality-specific features \( \textbf{F}_r \) and \( \textbf{F}_x \) from \( \textbf{I}_r \) and \( \textbf{I}_x \), respectively. These features are then refined using a \textbf{shared transformer encoder}, where self-attention is constrained within RGB and non-RGB modality to preserve modality-specific representations. Following prior works \cite{mask2former,maskformer}, we formulate semantic segmentation as a \textbf{\textit{mask classification task}}. The enhanced modality-specific features serve as keys and values in a shared transformer decoder, which processes them using modality-specific queries $\textbf{Q} = \{\textbf{Q}_r, \textbf{Q}_x\}$, where $\textbf{Q}_r, \textbf{Q}_x \in \mathbb{R}^{L \times C}$ denote the queries for the RGB and X modalities. Here, $L$ represents the number of queries per modality, and $C$ denotes the feature dimension. To refine $\textbf{Q}_r$ and $\textbf{Q}_x$, we apply modality-specific cross-attention:
\begin{equation}
\setlength{\abovedisplayskip}{10pt}
\setlength{\belowdisplayskip}{10pt}
\begin{aligned}
\textbf{Q}_r &= \text{softmax} \left( \frac{f_q(\textbf{Q}_r) f_k(\textbf{F}_r)^\top}{\sqrt{C}} \right) f_v(\textbf{F}_r), \\
\textbf{Q}_x &= \text{softmax} \left( \frac{f_q(\textbf{Q}_x) f_k(\textbf{F}_x)^\top}{\sqrt{C}} \right) f_v(\textbf{F}_x),
\end{aligned}
\end{equation}
where \( f_q \), \( f_k \), and \( f_v \) denote the projection for query, key, and value, respectively. Finally, the modality-specific queries \( \textbf{Q} \) are fed into a classifier to obtain the classification results $\textbf{C} = \{\textbf{C}_r, \textbf{C}_x\}$, where $\textbf{C}_r, \textbf{C}_x \in [0,1]^{L \times (K+1)}$ are the classification scores for RGB and X modalities, respectively. Here, \( K+1 \) accounts for \( K \) object classes and an additional ``None'' class (not matched). Simultaneously, the queries compute the inner product with the features from the mask encoder, generating modality-specific masks $\textbf{M} = \{\textbf{M}_r, \textbf{M}_x\}$, where $\textbf{M}_r, \textbf{M}_x \in [0,1]^{L \times H \times W}$ are the predicted segmentation masks from the RGB and X modalities. Finally, we construct the final prediction set $\textbf{P}$ by combining classification predictions $\textbf{C}$ and mask predictions $\textbf{M}$ at the query level. Formally, each prediction is represented as a tuple $(c^i, \textbf{m}^i)$, where $c^i \in \textbf{C}$ denotes the predicted class label, and $\textbf{m}^i \in \textbf{M}$ represents the corresponding mask prediction. The complete set of predictions is defined as $\textbf{P} = \{(c^i, \textbf{m}^i) \mid c^i \in \textbf{C}, \textbf{m}^i \in \textbf{M}\}_{i=0}^{2L}$. 



\subsection{Unified Modality Matching (UMM)}
To further enhance the contribution of RGB and non-RGB modality, we propose Unified Modality Matching (UMM), as illustrated in Fig.~\ref{fig:umm}. UMM consists of Modality Agnostic Matching (MAM) and Complementary Matching (CM), ensuring balanced label matching across RGB and X modalities. These steps must be executed sequentially, with MAM first for global matching, followed by CM to refine matching, ensuring optimality across modalities.

\begin{figure}[t]
\begin{center}
\includegraphics[width=1\linewidth]{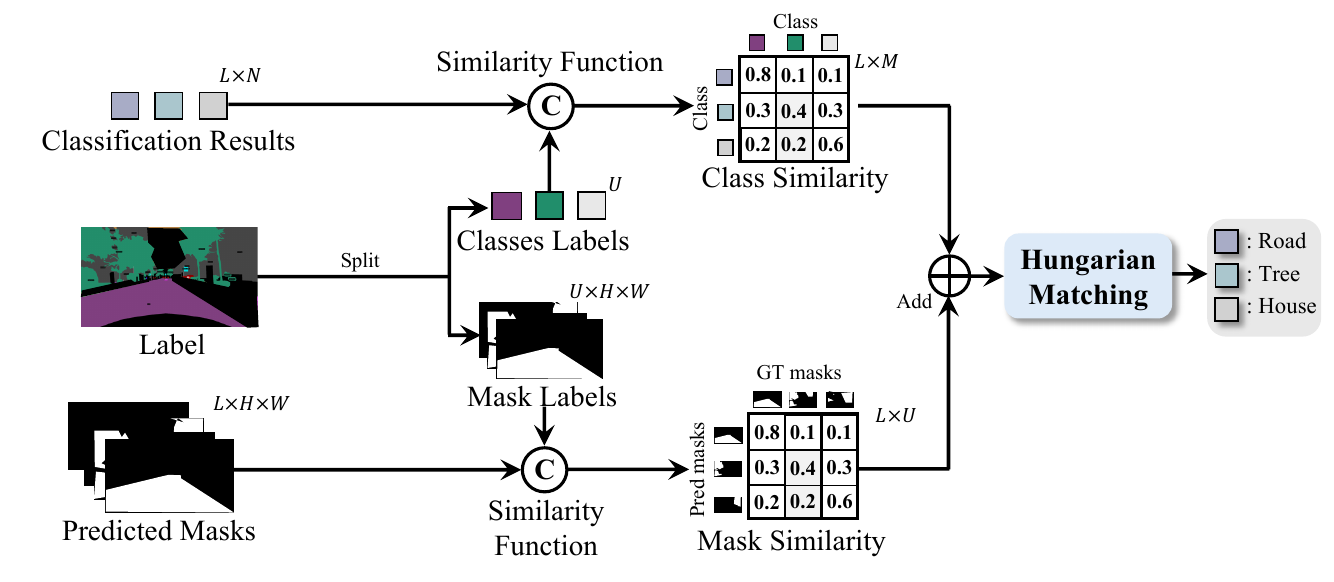}
\end{center}
\vspace{-0.2in}
\caption{How is hungarian matching applied in segmentation.}
\label{fig:hungarian}
\vspace{-0.2in}
\end{figure}

We begin with modality agnostic matching, where predictions from both RGB and X modalities are treated as a unified set. The goal of MAM is to assign labels to queries in a globally optimal manner, without considering modality constraints. The optimal matching is obtained by solving:
\begin{equation}
\setlength{\abovedisplayskip}{3pt}
\setlength{\belowdisplayskip}{3pt}
\sigma^a = \arg\min_{\sigma} \sum_{i=1}^{2L} \mathcal{L}_{match}(\textbf{P}^{\sigma(i)}, \textbf{Y}),
\end{equation}
where the matching loss is defined as:
\begin{equation}
\setlength{\abovedisplayskip}{5pt}
\setlength{\belowdisplayskip}{5pt}
\mathcal{L}_{match}(\textbf{P}, \textbf{Y}) = \mathcal{L}_{cls}(\textbf{C}, \textbf{Y}) + \mathcal{L}_{mask}(\textbf{M}, \textbf{Y}),
\end{equation}
where $\mathcal{L}_{cls}$ and $\mathcal{L}_{mask}$ indicate the loss function for classification and mask prediction. Class labels ranging from \( 1 \) to \( K \) represent valid classes, while \( 0 \) indicates the ``None'' category, meaning the prediction is not associated with any ground truth object. For such ``None'' queries, the corresponding mask is explicitly defined as $\mathbf{0}^{H \times W}$. The obtained matching is $\sigma^a = \big\{ (\textbf{p}^i, \textbf{y}_a^i) \big\}_{i=0}^{2L}$ where \( \textbf{p}^i \in \textbf{P} \) is the prediction of the \( i \)th query, and \( \textbf{y}_a^i \) is the assigned ground truth. For better understanding, the process of computing matching for segmentation is shown in Fig. \ref{fig:hungarian}.

Although this matching step is globally optimal across modalities, it does not account for modality balance, potentially leading to all labels being assigned to a single modality, resulting in many categories being unrecognized when a modality is missing. To this end, we introduce complementary matching. First, we decompose the initial matching \( \sigma^a \) into separate matching for RGB and non-RGB modality $\sigma^a_r = \{ (\textbf{p}^{i}_r, \textbf{y}_r^i) \ | \ \textbf{p}^{i}_r \in \textbf{P}_r \}, \quad
\sigma^a_x = \{ (\textbf{p}_{x}^i, \textbf{y}_x^i) \ | \ \textbf{p}_{x}^i \in \textbf{P}_x \}$ where \( \textbf{P}_r \) and \( \textbf{P}_x \) indicate the predictions from RGB and X modalities, respectively. 
By doing so, we extract the set of labels that have already been matched to queries in RGB and non-RGB modality $\textbf{Y}_r = \{ \textbf{y}_r^i \ | \ (\textbf{p}^i_r, \textbf{y}_r^i) \in \sigma^a_r \}, \quad
\textbf{Y}_x = \{ \textbf{y}_x^i \ | \ (\textbf{p}^{i}_x, \textbf{y}_x^i) \in \sigma^a_x \}$.
The remaining unassigned labels for RGB and non-RGB modality are computed as $\hat{\textbf{Y}}_r = \textbf{Y} \setminus \textbf{Y}_r, \quad
\hat{\textbf{Y}}_x = \textbf{Y} \setminus \textbf{Y}_x$ where $\setminus$ indicates the set minus.
Similarly, we define the set of unmatched queries as: $\hat{\textbf{P}}_r = \{\textbf{p}_{r}^i \ | \ C(\textbf{y}^i_r) = 0, (\textbf{p}_{r}^i, \textbf{y}^i_r) \in \sigma_r^a \}, \
\hat{\textbf{P}}_x = \{\textbf{p}_{x}^i \ | \ C(\textbf{y}^i_x) = 0, (\textbf{p}_{x}^i, \textbf{y}^i_x) \in \sigma_x^a \}$. We perform complementary matching within RGB and non-RGB modality:
\vspace{-8pt}
\begin{equation}
\begin{aligned}
\setlength{\abovedisplayskip}{3pt}
\setlength{\belowdisplayskip}{3pt}
    \sigma^s_r &= \arg\min_{\sigma} \sum_{i=1}^{\hat{L}_r} \mathcal{L}_{match}(\hat{\textbf{P}}^{\sigma(i)}_r, \hat{\textbf{Y}}_r), \\
    \sigma^s_x &= \arg\min_{\sigma} \sum_{i=1}^{\hat{L}_x} \mathcal{L}_{match}(\hat{\textbf{P}}^{\sigma(i)}_x, \hat{\textbf{Y}}_x),
\end{aligned}
\end{equation}
where \( \hat{L}_r \) and \( \hat{L}_x \) represent the number of remaining unassigned predictions in the RGB and X modalities.

\begin{algorithm}[t!]
\caption{Unified Modality Matching (UMM)}
\label{alg:umm}
\begin{algorithmic}[1]
\Require Prediction sets $\textbf{P}_r, \textbf{P}_x$ from RGB and X modalities
\Require Ground truth labels $\textbf{Y}$, matching loss $\mathcal{L}_{\text{match}}$
\Ensure Final matching $\sigma^*$

\State \textbf{Step 1: Modality Agnostic Matching (MAM)}
\State Construct prediction set $\textbf{P} = \{\textbf{p}^{i}_{r}, \textbf{p}^{i}_{x} \}_{i=1}^{L}$
\State Solve:
\[
\sigma^a = \arg\min_{\sigma} \sum_{i=1}^{2L} \mathcal{L}_{\text{match}}(\textbf{P}^{\sigma(i)}, \textbf{Y})
\]
\State Split $\sigma^a$ into complementary matching: $\sigma^a_r$ (RGB) and $\sigma^a_x$ (X)

\State \textbf{Step 2: Identify Unassigned Predictions}
\State Compute assigned labels:
\[
\textbf{Y}_r = \{ \textbf{y}_r^i \ | \ (\textbf{p}^i_r, \textbf{y}_r^i) \in \sigma^a_r \}, \quad
\textbf{Y}_x = \{ \textbf{y}_x^i \ | \ (\textbf{p}^i_x, \textbf{y}_x^i) \in \sigma^a_x \}
\]
\State Compute unassigned labels:
\[
\hat{\textbf{Y}}_r = \textbf{Y} \setminus \textbf{Y}_r, \quad
\hat{\textbf{Y}}_x = \textbf{Y} \setminus \textbf{Y}_x
\]
\State Compute unassigned predictions:
\[
\hat{\textbf{P}}_r = \{\textbf{p}_{r}^i \ | \ C(\textbf{y}^i_r) = 0, (\textbf{p}_{r}^i, \textbf{y}^i_r) \in \sigma_r^a \}
\]
\[
\hat{\textbf{P}}_x = \{\textbf{p}_{x}^i \ | \ C(\textbf{y}^i_x) = 0, (\textbf{p}_{x}^i, \textbf{y}^i_x) \in \sigma_x^a \}
\]

\State \textbf{Step 3: Complementary Matching (CM)}
\State Solve:
\[
\sigma^s_r = \arg\min_{\sigma} \sum_{i=1}^{\hat{L}_r} \mathcal{L}_{\text{match}}(\hat{\textbf{P}}^{\sigma(i)}_r, \hat{\textbf{Y}}_r)
\]
\[
\sigma^s_x = \arg\min_{\sigma} \sum_{i=1}^{\hat{L}_x} \mathcal{L}_{\text{match}}(\hat{\textbf{P}}^{\sigma(i)}_x, \hat{\textbf{Y}}_x)
\]

\State \textbf{Step 4: Merge matchings}
\State Extend $\sigma^s_r$ and $\sigma^s_x$ to match the size of $\sigma^a$ by adding ``None'' placeholders
\State For predictions already assigned in $\sigma^a$, set their corresponding entries in $\sigma^s_r$ and $\sigma^s_x$ to ``None''
\State Compute the final label matching:
\[
\textbf{y}^i_{*} = (\max(c^i_a, c^i_{s_r}, c^i_{s_x}), \ \max(\textbf{m}^i_a, \textbf{m}^i_{s_r}, \textbf{m}^i_{s_x}))
\]
\State Obtain final matching:
\[
\sigma^* = \sigma^a \cup \sigma^s_r \cup \sigma^s_x
\]

\State \textbf{Step 5: Compute Segmentation Loss}
\State Compute final loss:
\[
\mathcal{L}_{seg}(\textbf{P}, \textbf{Y}) = \mathcal{L}_{cls}(\textbf{C}^{\sigma^*}, \textbf{Y}) + \mathcal{L}_{mask}(\textbf{M}^{\sigma^*}, \textbf{Y})
\]
\State \Return $\sigma^*$
\end{algorithmic}
\end{algorithm}

To unify both matching results into a single one, we integrate modality agnostic and modality specific matches. Concretely, to merge the results from modality agnostic and complementary matching into a unified matching $\sigma^*$, we first extend $\sigma^s_r$ and $\sigma^s_x$ to the same length as $\sigma^a$. This is done by ensuring that each query in  $\sigma^s_r$ has a corresponding placeholder in $\sigma^s_x$  set to ``None’’ and vice versa.  Next, for queries already assigned in $\sigma^a$, we set their corresponding entries in $\sigma^s_r$ and $\sigma^s_x$ to ``None’’. The final matching result is $\sigma^* = \{(\textbf{p}^i, \textbf{y}^i_{*}) \mid \textbf{p}^i \in \textbf{P}\}$ where each prediction \( \textbf{p}^i \) is assigned the final label \( \textbf{y}^i_{*} \). The final label matching is computed as $\textbf{y}^i_{*} = (\max(c^i_a,c^i_{s_r},c^i_{s_x}), \ \max(\textbf{m}^i_a,\textbf{m}^i_{s_r},\textbf{m}^i_{s_x}))$
where \( (c^i_{a}, \textbf{m}^i_{a}) \), \( (c^i_{s_r}, \textbf{m}^i_{s_r}) \), and \( (c^i_{s_x}, \textbf{m}^i_{s_x}) \) denote the class and mask matching from \( \sigma^a \), \( \sigma^s_r \), and \( \sigma^s_x \), respectively. The final segmentation loss is:
\begin{equation}
\mathcal{L}_{seg}(\textbf{P}, \textbf{Y}) = \mathcal{L}_{cls}(\textbf{C}^{\sigma^*}, \textbf{Y}) + \mathcal{L}_{mask}(\textbf{M}^{\sigma^*}, \textbf{Y}),
\end{equation}
where $\textbf{C}^{\sigma^*}$ and $\textbf{M}^{\sigma^*}$ indicate the class and mask prediction under the matching ${\sigma^*}$. The pseudo code of UMM is shown in Algorithm \ref{alg:umm} for better understanding.

\begin{figure}[t]
\begin{center}
\includegraphics[width=1\linewidth]{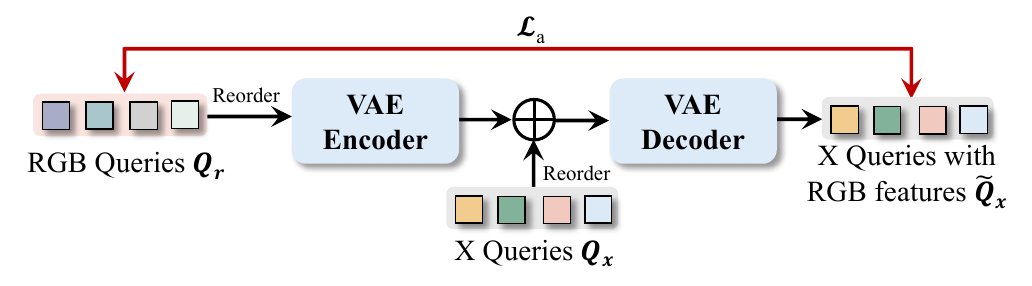}
\end{center}
\vspace{-0.2in}
\caption{Overview of cross modality alignment where we only show the process to refine RGB queries for simplicity.}
\label{fig:cross}
\vspace{-0.2in}
\end{figure}

\begin{table*}[t!]
\caption{Comparison with previous methods on real-world benchmark MUSES \cite{muses} with three modalities (F: frame camera, E: event camera, L: LiDAR sensor). \textbf{Bold} indicates the highest performance while \underline{underline} indicates the second best performance. Mean refers to the average mIoU across all modality settings.}
\vspace{-0.1in}
\setlength{\tabcolsep}{8pt}
\resizebox{\linewidth}{!}{

\begin{tabular}{c|c|c|c|c|ccccccc|c}
\toprule
\multirow{2}{*}{Method} & \multirow{2}{*}{Backbone} & \multirow{2}{*}{\begin{tabular}[c]{@{}c@{}}Training\\ Modality\end{tabular}} & \multirow{2}{*}{Parameters (M)$\downarrow$} & \multirow{2}{*}{GFLOPs $\downarrow$} & \multicolumn{7}{c|}{Testing Modality}                                                                                        & \multirow{2}{*}{Mean}  \\ \cmidrule{6-12}
                        &                           &                                                                              &                                 &                         & F              & E              & L              & FE             & FL             & EL             & FEL            &                        \\ \midrule
CMNeXt \cite{deliver}                 & Seg-B0  \cite{segformer}                  & F                                                                            & \textbf{3.72}                               & \textbf{27.46}                       & 43.37          & -              & -              & -              & -              & -              & -              & -                      \\ \midrule
CMX \cite{cmx}                     & \multirow{4}{*}{Seg-B0 \cite{segformer}}   & \multirow{4}{*}{FEL}                                                         & 10.32                               & 62.40                       & 2.52           & 2.35           & 3.01           & 41.15          & 41.25          & 2.55           & 19.30          & 16.10                  \\
CMNeXt \cite{deliver}                  &                           &                                                                              & {\ul 10.30}                               & {\ul 44.78}                       & 3.50           & 2.77           & 2.64           & 6.63           & 10.28          & 3.14           & 46.66          & 10.80                  \\
MAGIC \cite{magic}                  &                           &                                                                              & 24.74                               & 444.35                       & 43.22          & 2.68           & 22.95          & 43.51          & 49.05          & 22.98          & 49.02          & 33.34                  \\
Any2Seg \cite{any2seg}                &                           &                                                                              & 24.73                               & -                       & 44.40    &  3.17     & 22.33    & 44.51    &  49.96    & 22.63    &  50.00    & 33.86            \\ \midrule
\multirow{4}{*}{Ours}   & ResNet-18 \cite{resnet}                 & \multirow{4}{*}{FEL}                                                         & 28.91                     & 53.54                   & 62.03          & 17.20          & 39.33          & 57.27          & 59.56          & 45.76          & 62.40          & 49.07 (15.21$\uparrow$)          \\
                        & ResNet-34 \cite{resnet}              &                                                                              & 53.75                           & 91.85                   & {\ul 66.70}          & {\ul 20.60}          & \textbf{46.89} & {\ul 61.14}          & {\ul 68.22}          & \textbf{54.76} & {\ul 69.46}          & {\ul 55.39 (21.53$\uparrow$)}          \\
                        & ResNet-50 \cite{resnet}                &                                                                              & 58.94                           & 100.0                   & \textbf{70.10} & \textbf{22.18} & {\ul 45.57}          & \textbf{64.59} & \textbf{69.28} & {\ul 52.44}          & \textbf{71.36} &  \textbf{56.50 (22.74$\uparrow$)}    \\ \cmidrule{2-2} \cmidrule{4-13} 
                        & Seg-B0 \cite{segformer}                    &                                                                              & 17.70                           & 59.10                   & 65.76          &    16.88       & 39.63          & 58.42          & 59.77          & 49.89          & 65.81          & 50.88 (17.02$\uparrow$) \\ \bottomrule
\end{tabular}
}
\label{tab:sota muses}
\vspace{-0.05in}
\end{table*}

\begin{table*}[t!]
\caption{Comparison with previous methods on synthetic benchmark DELIVER \cite{deliver} with four modalities (R: RGB images, D: depth images, E: event Camera, L: Lidar sensor) using ResNet-34 \cite{resnet} as backbone model. \textbf{Bold} indicates the highest performance while \underline{underline} indicates the second best performance. Mean refers to the average mIoU across all modalitys.}
\vspace{-0.1in}
\setlength{\tabcolsep}{4pt}
\resizebox{\linewidth}{!}{
\begin{tabular}{c|ccccccccccccccc|c}
\toprule
\multirow{2}{*}{Method} & \multicolumn{15}{c|}{Testing Modality}                                                                                     & \multirow{2}{*}{Mean} \\ \cmidrule{2-16}
                        & R     & D     & E    & L    & RD    & RE    & RL    & DE    & DL    & EL   & RDE   & RDL   & REL   & DEL   & RDEL  &                       \\ \midrule
CMNeXt \cite{deliver}                  & 0.86  & 0.49  & 0.66 & 0.37 & 47.06 & 9.97  & 13.75 & 2.63  & 1.73  & 2.85 & 59.03 & 59.18 & 14.73 & 59.18 & 39.07 & 20.77                 \\
MAGIC \cite{magic}                  & \underline{32.60} & \textbf{55.06} & \underline{0.52} & \underline{0.39} & \textbf{63.32} & \underline{33.02} & \underline{33.12} & \textbf{55.16} & \textbf{55.17} & \underline{0.26} & \textbf{63.37} & \textbf{63.36} & \underline{33.32} & \textbf{55.26} & \textbf{63.40} & \underline{40.49}                 \\ \midrule
Ours                    & \textbf{53.32 }& \underline{50.18} & \textbf{1.03} & \textbf{1.49} & \underline{55.58} & \textbf{53.30} & \textbf{53.40} & \underline{49.60} & \underline{51.20} & \textbf{2.19} & \underline{55.57} & \underline{55.70} & \textbf{53.77} & \underline{54.08} & \underline{58.29} & \textbf{43.24} (\textbf{2.75$\uparrow$})                \\ \bottomrule
\end{tabular}
}
\label{tab:sota deliver}
\vspace{-0.2in}
\end{table*}

\subsection{Cross Modality Alignment (CMA)}
Though effective, the proposed Unified Modality Matching (UMM) framework may result in suboptimal performance due to its query assignment strategy. Specifically, for one category, UMM assigns high-quality queries to one modality while using less effective queries in the other, leading to degraded predictions when combining both modalities. To address this, we propose Cross Modality Alignment (CMA), as illustrated in Fig.~\ref{fig:cross}. CMA leverages the most discriminative queries assigned in MAM to enhance their weaker counterparts in CM, ensuring that both modalities utilize high-quality queries for all categories. By aligning weaker queries with their stronger counterparts across modalities, CMA enables more robust feature learning and improves overall segmentation performance.

Given the query sets $\textbf{Q}_r$ and $\textbf{Q}_x$, along with the final matching results $\sigma^*_r$ and $\sigma^*_x$, where $\sigma^*_r = \{ (\textbf{p}^i, \textbf{y}_*^i) \mid \textbf{p}^i \in \textbf{P}_r \}, \quad
\sigma^*_x = \{ (\textbf{p}^i, \textbf{y}_*^i) \mid \textbf{p}^i \in \textbf{P}_x \}$, we first reorder queries based on their assigned class labels, ensuring a consistent indexing structure across modalities. Due to inherent modality differences, directly aligning the queries across modalities may reduce modality discrepancies rather than preserving semantic consistency. To address this, we then introduce a VAE-based refiner to facilitate cross-modal adaptation while explicitly controlling modality discrepancy and semantic alignment. We choose VAE as the refinement model because, unlike GANs, it does not require additional adversarial training steps, making it more stable and efficient. While diffusion models have shown strong generative capabilities, they typically require the target distribution to be relatively static for effective generation. However, in our method, both the generated target and the input are dynamically changing, making diffusion-based approaches less suitable for our setting. Specifically, the encoder processes RGB queries to extract modality-specific characteristics and infuse them into the X queries, ensuring that essential modality distinctions are retained rather than collapsed. Meanwhile, the decoder refines X query representations by minimizing the discrepancy between modality-specific class centers and their shared semantic counterparts, ensuring that queries from different modalities converge toward the same category representation. The alignment loss is:
\begin{equation}
    \mathcal{L}_{a} = \mathcal{L}_{r}(\textbf{Q}_r,\tilde{\textbf{Q}}_x) + \mathcal{L}_{r}(\textbf{Q}_x,\tilde{\textbf{Q}}_r),
\end{equation}
where \( \mathcal{L}_r \) is a combination of Mean Squared Error (MSE) and Maximum Mean Discrepancy (MMD) loss \cite{pading,concat}. 


\begin{table}[t]
\setlength{\tabcolsep}{4pt}
\caption{Comparison with previous methods on NYUv2 \cite{nyuv2} with two modalities (RGB and depth) \textbf{Bold} indicates the highest performance while \underline{underline} indicates the second best performance. The number in () indicates the performance without multi-scale inference.}
\resizebox{\linewidth}{!}{
\begin{tabular}{cccc}
\toprule
Model       & Backbone    & One / Two Stage & mIoU                \\ \midrule
ACNet \cite{acnet}      & ResNet-50 \cite{resnet}   & One             & 48.3                \\
ShapeConv \cite{shapeconv}  & ResNext-101 \cite{resnext} & One             & 51.3                \\
ESANet \cite{esanet}     & ResNet-34 \cite{resnet}  & One             & 50.3                \\
TokenFusion \cite{tokenfusion} & MiT-B3 \cite{segformer}     & One             & 54.2                \\
Omnivore \cite{omnivore}   & Swin-B \cite{swin}      & One             & 54.0                \\
CMNext \cite{deliver}     & MiT-B4 \cite{segformer}     & One             & { \ul 56.9} ({\ul 53.6})          \\
GOPT \cite{aaaifusion1}       & ViT \cite{vit}        & One             & 54.3       \\
Dformer \cite{dformer}    & Dformer-L \cite{dformer}   & Two             & \textbf{57.2}                \\
Ours        & MiT-B4 \cite{segformer}      & One             & 54.5 (\textbf{54.1}) \\ \bottomrule
\end{tabular}
}
\label{tab:nyu}
\vspace{-0.2in}
\end{table}

\subsection{Training Objectives and Inference}

The total loss functions for training are defined as,
\begin{equation}
\setlength{\abovedisplayskip}{3pt}
\setlength{\belowdisplayskip}{3pt}
    \mathcal{L} = \mathcal{L}_{seg} +\mathcal{L}_{a}.
\end{equation}
During inference, modality-specific queries from the backbone and transformer encoder are passed to the classifier. The classification results are then combined with the predicted masks via the inner product for the prediction. If any sub-modality in X is missing, it is replaced with $\textbf{0}^{3 \times H \times W}$.

\begin{table*}[t!]
\caption{Performance of each classes on Muses \cite{muses}.}
\vspace{-0.1in}
\setlength{\tabcolsep}{8pt}
\resizebox{\linewidth}{!}{
\begin{tabular}{ccccccccccc}
\toprule
\multicolumn{1}{c|}{Class} & Road  & Sidewalk & Building & Wall  & Fence & Pole  & Traffic Light & Traffic Sign & Vegetation & Terrain      \\ \midrule
\multicolumn{1}{c|}{mIoU}  & 96.58 & 80.08    & 89.41    & 57.66 & 57.92 & 56.39 & 69.11         & 67.73        & 86.28      & 73.28        \\ \midrule \midrule
\multicolumn{1}{c|}{Class} & Sky   & Person   & Rider    & Car   & Truck & Bus   & Train         & Motorcycle   & Bicycle    & Average mIoU \\ \midrule
\multicolumn{1}{c|}{mIoU}  & 94.05 & 62.80    & 39.96    & 88.45 & 52.81 & 58.01 & 96.96         & 43.0         & 49.31      & 69.46        \\ \bottomrule
\end{tabular}
}
\label{tab:class performance}
\vspace{-0.2in}
\end{table*}

\section{Experiments}
\label{sec:experiments}
\begin{table}[t!]
\caption{Ablation studies on proposed methods where ``O'' ,``T'', ``U'', ``A'' indicate one shared backbone, two separate backbones, UMM, and CMA, respectively.}
\vspace{-0.05in}
\renewcommand{\tabcolsep}{2.5pt}
\resizebox{\linewidth}{!}{
\begin{tabular}{c|ccccccc|c}
\toprule
\multirow{2}{*}{Method} & \multicolumn{7}{c|}{Testing Modality}                         & \multirow{2}{*}{Mean} \\ \cmidrule{2-8}
                        & F     & E     & L     & FE    & FL    & EL    & FEL   &                       \\ \midrule
O           & 24.77     & 2.99     & 1.79     & 24.05     & 22.65     & 2.82     & 24.68     & 14.82                     \\
O + U            & 34.28     & 20.30     & 21.34     & 35.71     & 36.64     & 38.10     & 43.30     & 32.66                     \\
T + U                  & \textbf{68.35} & \textbf{21.23} & 45.63 & 60.89 & 66.83 & 50.68 & 67.70 & 54.47                 \\
T + U + A              & 66.70 & 20.60 & \textbf{46.89} & \textbf{61.14} & \textbf{68.22} & \textbf{54.76} & \textbf{69.46} & \textbf{55.39}                  \\ \bottomrule
\end{tabular}
}
\label{tab:ablation}
\vspace{-0.1in}
\end{table}

\subsection{Experimental Setup}
\noindent \textbf{Datasets.} We evaluate our method on both synthetic and real-world multi-sensor datasets to assess its robustness and generalizability. The MUSES dataset \cite{muses} is a real-world driving dataset collected in Switzerland, specifically designed to address challenges arising from adverse visual conditions. It provides multi-sensor recordings from a high-resolution RGB camera, an event camera, and a MEMS LiDAR, along with high-quality 2D panoptic annotations for benchmarking. The DELIVER \cite{deliver} is a synthetic dataset which comprises RGB, depth, LiDAR, and event streams across 25 semantic categories. It encompasses a wide range of environments and includes scenarios involving sensor degradation and failure, making it well-suited for evaluating multimodal robustness. NYUv2 Dataset. NYUv2 is a standard benchmark for indoor RGB-D semantic segmentation, consisting of 1,449 RGB-D images annotated with 40 semantic categories. The images are captured in diverse indoor scenes using a Kinect sensor. We follow the official split with 795 training and 654 testing images.

\noindent \textbf{Implementation Details.} All experiments were conducted on 4 NVIDIA A6000 GPUs, The initial learning rate was set to \(6 \times 10^{-5}\) and adjusted using a polynomial decay strategy with a power of 0.9 over 200 epochs. Additionally, a 10-epoch warm-up phase was applied at 10\% of the initial learning rate to stabilize training. AdamW optimizer was employed, and the effective batch size for both datasets was set to 16. Input modality data was cropped to \(1024 \times 1024\) across benchmarks. The VAE structures for both the encoder and decoder are four layers of Linear-Layernorm-ReLU.

\begin{table}[t]
\caption{Ablation studies on UMM.}
\vspace{-0.1in}
\renewcommand{\tabcolsep}{2pt}
\resizebox{\linewidth}{!}{
\begin{tabular}{cc|c|ccccccc|c}
\toprule
\multicolumn{2}{c|}{Method} & \multirow{2}{*}{GFLOPs$\downarrow$} & \multicolumn{7}{c|}{Testing Modality}                                                                                        & \multirow{2}{*}{Mean} \\ \cmidrule{4-10}
MAM          & CM          &                         & F              & E              & L              & FE             & FL             & EL             & FEL            &                       \\ \midrule
$\checkmark$          &              & 91.83                   & 55.17          & 6.09           & 6.62           & 47.02          & 55.09          & 7.74           & 55.50          & 33.31                 \\
             & $\checkmark$          & 91.85                   & 66.66          & 3.32           & \textbf{47.02} & 50.75          & 66.29          & 35.62          & 68.44          & 48.30                 \\
$\checkmark$          & $\checkmark$          & 91.85                   & \textbf{66.70} & \textbf{20.60} & 46.89          & \textbf{61.14} & \textbf{68.22} & \textbf{54.76} & \textbf{69.46} & \textbf{55.39}        \\ \bottomrule
\end{tabular}
}
\label{tab:ablation on matching}
\vspace{-0.2in}
\end{table}

\subsection{Comparison with Existing Methods}

We compare our method with previous approaches on MUSES (real-world) \cite{muses} and DELIVER (synthetic) \cite{deliver} datasets, including missing and full modality compared with fusion-based methods CMNeXt \cite{deliver} and knowledge-distillation based methods MAGIC \cite{magic} during testing. Table \ref{tab:sota muses} compares our approach with fusion-based and knowledge distillation (KD)-based methods on the MUSES benchmark \cite{muses}. Feature fusion methods, such as CMNeXt \cite{deliver}, integrate multi-modal features at the representation level but suffer significant performance drops when trained with only RGB. Additionally, CMX \cite{cmx} and CMNeXt \cite{deliver} exhibit severe degradation when individual modalities are used in isolation, demonstrating their sensitivity to missing modalities. In contrast, KD-based methods, such as MAGIC \cite{magic}, rely on teacher models and achieve strong multi-modal performance (\textbf{49.02} mIoU) but suffer from excessive computational costs. Our approach overcomes these limitations by enabling robust multi-modal learning with reduced computational overhead. Unlike fusion models, our framework remains effective even when certain modalities are missing. 

On DELIVER as shown in Table \ref{tab:sota deliver}, our approach achieves the best results in single-modality settings, particularly excelling in RGB (53.32\%) and event-based segmentation (1.03\%), while maintaining competitive performance in depth and LiDAR. Notably, MAGIC achieves slightly higher mIoU when all modalities are present. However, in mixed-modality settings, MAGIC’s performance collapses if a modality is missing (mean 40.49\%), whereas our method maintains strong results. 

As shown in Table~\ref{tab:nyu}, our method achieves competitive performance on the NYUv2 dataset under the RGB-D setting, reaching 54.5 mIoU, and outperforming most existing single-stage approaches. Importantly, unlike DFormer~\cite{dformer}, which attains the highest accuracy but relies on a two-stage distillation pipeline with training an additional teacher model, our method is entirely one-stage and training-efficient, requiring no auxiliary supervision or extra models. Moreover, our performance remains stable even without multi-scale inference (54.5 → 54.1), in stark contrast to methods like CMNeXt, which exhibit a substantial performance drop under the same setting (56.9 → 53.6). This result underscores the inherent robustness and efficiency of our approach, making it better suited for practical deployment scenarios where test-time augmentation is limited.

\begin{table}[t!]
\caption{Ablation studies on the refiner in CMA, where “MLP”, ``M'', ``C'', “Align” denote MLP-based refiner, modality distance, class distance, align both modalities, respectively.}
\vspace{-0.1in}
\renewcommand{\tabcolsep}{2pt}
\resizebox{\linewidth}{!}{
\begin{tabular}{c|c|c|ccccccc|c}
\toprule
\multirow{2}{*}{Method} & \multirow{2}{*}{M$\uparrow$} & \multirow{2}{*}{C$\downarrow$} & \multicolumn{7}{c|}{Testing Modality}                                                                                        & \multirow{2}{*}{Mean} \\ \cmidrule{4-10}
                        &                      &                     & F              & E              & L              & FE             & FL             & EL             & FEL            &                       \\ \midrule
Align              & 0.27                 & 2.76                & 64.54          & \textbf{21.51} & 44.07          & 59.21          & 62.66          & 50.27          & 63.79          & 52.29                 \\
MLP                     & 0.93                 & 0.57                & 65.32          & 20.64          & 43.47          & 61.09          & 62.18          & 52.13          & 65.16          & 52.85                 \\
VAE              & 1.03                 & 0.49                & \textbf{66.70} & 20.60          & \textbf{46.89} & \textbf{61.14} & \textbf{68.22} & \textbf{54.76} & \textbf{69.46} & \textbf{55.39}        \\ \bottomrule
\end{tabular}
}
\label{tab:refine}
\vspace{-0.2in}
\end{table}

\noindent \textbf{Per class performance on MUSES \cite{muses}.} Table \ref{tab:class performance} presents the per-class mIoU performance of our method on the MUSES benchmark. Our model achieves high accuracy on well-defined and dominant classes such as Road, Building, and Sky, where clear structural boundaries and large-scale presence facilitate robust segmentation. Similarly, classes with distinct shapes and textures, such as Train and Vegetation, also exhibit strong performance. However, performance degrades on more challenging categories, particularly small or occluded objects. Thin structures like Pole and Traffic Light  exhibit moderate accuracy, likely due to their limited spatial footprint. Human-related classes, such as Person  and Rider, as well as vehicles with complex geometry, including Truck and Motorcycle, achieve relatively lower mIoU, indicating challenges in fine-grained segmentation and boundary preservation. Despite these challenges, our method maintains a competitive 69.46\% average mIoU across all classes, demonstrating strong generalization capabilities while highlighting areas for future improvements, particularly in small-object recognition and occlusion handling.

\subsection{Ablation Studies}

To evaluate the effectiveness of the proposed methods, we conduct ablation studies on the MUSES dataset with 200 epochs, using ResNet-34 as the backbone while keeping all hyperparameters unchanged.

\noindent \textbf{Ablations on the proposed modules.} Table \ref{tab:ablation} presents the ablation study on the proposed methods. Utilizing a single backbone without additional enhancements achieves the lowest performance with a mean mIoU of 14.82\%, highlighting its limited capability in multimodal settings. Adding UMM (O + U) significantly improves segmentation across all modalities, achieving a mean mIoU of 32.66\%, demonstrating the effectiveness of unified modality matching. When employing two backbones with UMM (T + U), the performance further improves, particularly in the F, FL, and FEL settings, yielding an overall mean mIoU of 54.47\%. Finally, integrating CMA (T + U + R) leads to the best results, achieving 68.22\% on FL, 69.46\% on FEL, and an overall mean mIoU of 55.39\%, indicating that the combination of multiple backbones, modality matching, and cross-modality attention results in the best.

\noindent \textbf{Ablations on the unified modality matching.} Table \ref{tab:ablation on matching}  evaluates Unified Modality Matching (UMM) by analyzing its two components: Modality Agnostic Matching (MAM) and Complementary Matching (CM). The GFLOPs column shows that UMM introduces negligible computational overhead. MAM alone performs poorly (33.31\% mIoU) due to the lack of modality-specific alignment, while CM improves performance (48.30\% mIoU) but remains suboptimal. Combining both (UMM) achieves the best result with only a 0.02 GFLOPs increase, proving that modality-specific alignment is effective and efficient. Our two-step matching ensures that no single modality dominates all labels, even when it could predict them correctly. While this reduces image optimality, it enhances robustness by ensuring RGB and non-RGB modality contributes. 

\begin{figure}[t!]
\begin{center}
\includegraphics[width=0.95\linewidth]{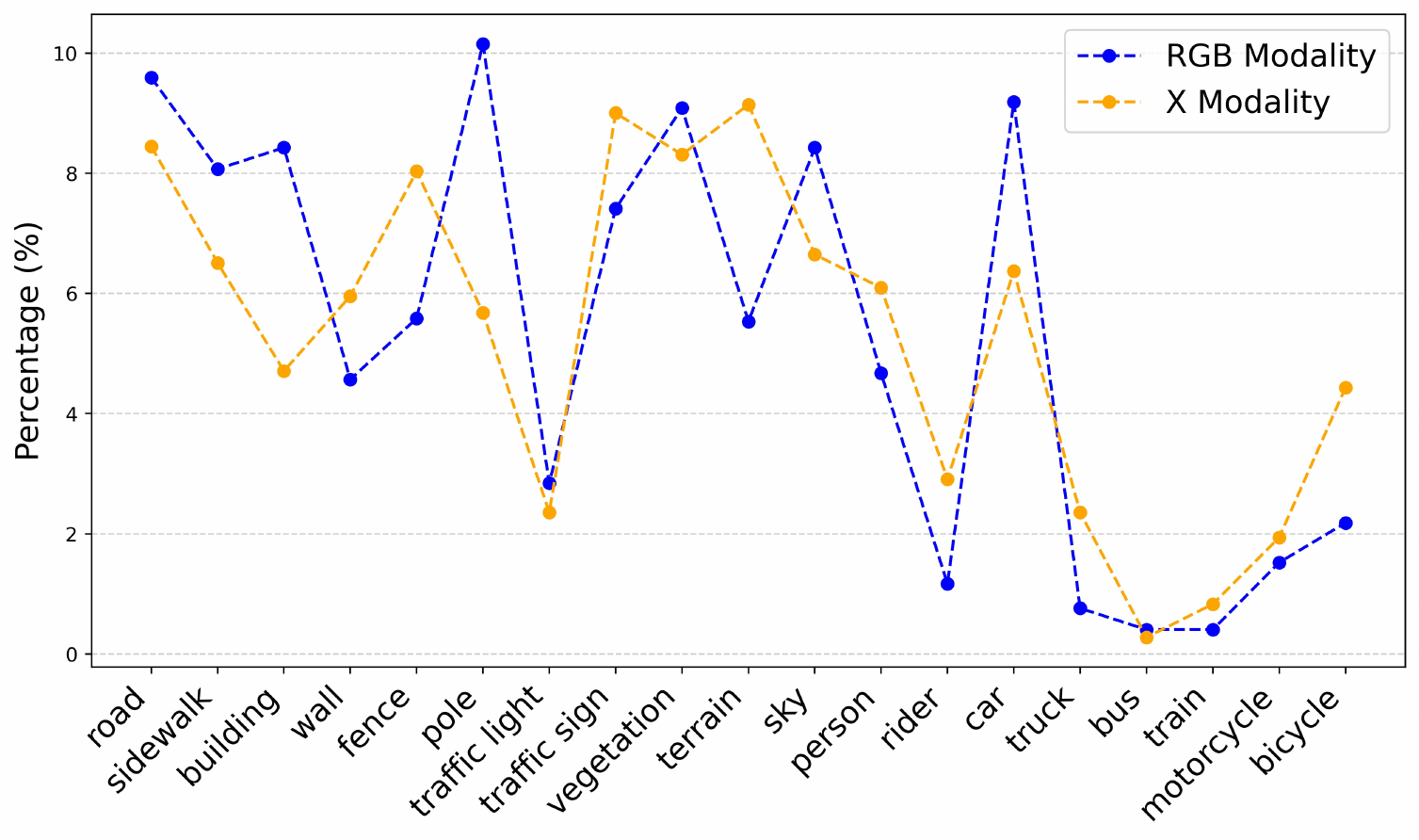}
\end{center}
\vspace{-0.2in}
\caption{Label matching distribution under UMM.}
\label{fig:distribution}
\vspace{-0.2in}
\end{figure}

To further analyze UMM’s label-matching behavior, we visualize the label distribution across semantic categories in the MUSES dataset, as shown in Fig. \ref{fig:distribution}. The x-axis represents category labels, while the y-axis denotes the percentage of labels assigned to RGB and non-RGB modality. The blue and orange dashed lines correspond to RGB and X modality matching, respectively. The results reveal that label matching is category-dependent, with fence and traffic signs favoring the X modality, while road and pole are predominantly assigned to RGB. This suggests that UMM effectively leverages the strengths of RGB and non-RGB modality rather than enforcing uniform matching. Additionally, categories such as vegetation exhibit a more balanced distribution, suggesting that both modalities predict the class they excel at. This adaptive matching highlights UMM’s ability to dynamically allocate queries based on modality-specific strengths. 

\begin{table}[t!]
\caption{Ablation studies on the number of backbones where ``S'' indicates using modality-specific backbones (3 backbones), and ``T'' indicates two backbones}
\setlength{\tabcolsep}{2pt}
\vspace{-0.1in}
\resizebox{\linewidth}{!}{
\begin{tabular}{c|c|c|ccccccc|c}
\toprule
\multirow{2}{*}{Method} & \multirow{2}{*}{Para(M)} & \multirow{2}{*}{GFLOPs$\downarrow$} & \multicolumn{7}{c|}{Testing Modality}                                                                                        & \multirow{2}{*}{Mean} \\ \cmidrule{4-10}
                        &                          &                         & F              & E              & L              & FE             & FL             & EL             & FEL            &                       \\ \midrule
``S''                  & 72.18                    & 138.0                   & 66.05          & \textbf{40.86} & 40.82          & \textbf{62.81}          & 66.05          & 41.95          & 52.39          & 52.99                 \\
``T''                    & \textbf{53.75}                    & \textbf{91.85}                   & \textbf{66.70} & 20.60 & \textbf{46.89} & 61.14 & \textbf{68.22} & \textbf{54.76} & \textbf{69.46} & \textbf{55.39}                  \\ \bottomrule
\end{tabular}
}
\label{tab:three modality}
\vspace{-0.1in}
\end{table}

\noindent \textbf{Ablation studies on the refiner in CMA. }
In multimodal segmentation, we identify two essential distance metrics: \textbf{modality distance} and \textbf{class distance}. \textbf{Modality distance} quantifies the distributional divergence between the two modalities, measuring their separability, while \textbf{class distance} represents the deviation of RGB and non-RGB modality’s features from their shared semantic center. Ideally, a well-balanced model should preserve modality-specific information while ensuring consistent category-wise representations. Our approach aims to maximize the contribution of RGB and non-RGB modality by \textbf{increasing modality distance} to retain distinct modality representations while \textbf{minimizing class distance}. 

To validate the hypothesis, Table \ref{tab:refine} evaluates refinement strategies in Cross Modality Alignment (CMA). MMD \cite{concat,pading} is used to measure cross-modal separation, while L1 loss is used to assess intra-class consistency. Without a refiner, direct query alignment reduces modality differences (low MMD) without effectively improving category-wise consistency (high L1). This suggests that naive alignment forces different modalities into a shared representation space but fails to preserve their intrinsic characteristics, leading to suboptimal feature adaptation. The MLP refiner partially mitigates this issue by improving both modality separation (higher MMD) and intra-class compactness (lower L1), but the rigid transformation remains insufficient for capturing complex modality-specific variations. In contrast, our VAE refiner achieves the best trade-off by leveraging a learned latent space that preserves modality-specific distinctions while refining category-level consistency. This results in both a higher modality distance and improved intra-class compactness, leading to the best segmentation performance. These findings highlight that naive query alignment can weaken modality distinction instead of enhancing category-level coherence.

\noindent \textbf{Ablations on the modality specific backbones. }Table \ref{tab:three modality} presents an ablation study on the number of backbones used in our method. “S” represents the use of modality-specific backbones, where each modality has its dedicated backbone (totaling three backbones). In contrast, “T” refers to our approach, which utilizes only two backbones while maintaining strong multimodal representations. Despite utilizing fewer parameters and significantly lower computational costs, our method achieves higher performance. 

\begin{table}[t!]
\caption{Ablation studies on backbone structure where ``18'', ``34'' indicate the ResNet18 and ResNet34, respectively.}
\setlength{\tabcolsep}{4pt}
\vspace{-0.1in}
\resizebox{\linewidth}{!}{
\begin{tabular}{cc|ccccccc|c}
\toprule
\multicolumn{2}{c|}{Method} & \multicolumn{7}{c|}{Testing Modality}                                                                                       & \multirow{2}{*}{Mean} \\ \cmidrule{3-9}
RGB          & X            & F              & E             & L              & FE             & FL             & EL             & FEL            &                       \\ \midrule
18     & 34     & 65.29          & 20.1          & 45.94          & 63.70          & 65.80          & 56.52          & 68.50          & 55.12                 \\
34     & 18     & 66.30          & 20.4          & 44.38          & 64.04          & 70.07          & 52.79          & 68.87          & 55.26                 \\
34     & 34     &  \textbf{66.70} & 20.60 & \textbf{46.89} & \textbf{61.14} & \textbf{68.22} & \textbf{54.76} & \textbf{69.46} & \textbf{55.39}        \\ \bottomrule
\end{tabular}}
\vspace{-0.1in}
\label{tab:structure}
\end{table}


\noindent \textbf{Ablations on the structures of two backbones.} \textbf{Ablation Study on Backbone Selection.}  
Table \ref{tab:structure} presents an ablation study on different backbone configurations, where "18" and "34" denote ResNet-18 and ResNet-34, respectively. The study explores the impact of backbone choices for RGB and X modalities on segmentation performance across different modality settings. The results indicate that using a stronger backbone for the X modality improves performance across most settings. For example, when using ResNet-34 for X while keeping ResNet-18 for RGB, the mean mIoU increases from 55.12\% to 55.26\%. When both RGB and X use ResNet-34, the model achieves the best performance (55.39\% mean mIoU), with noticeable improvements in L (46.89\%), FE (61.14\%), and FL (68.22\%) settings, suggesting that increasing the backbone capacity for the X modality enhances overall segmentation performance, particularly in multi-modal scenarios.

\begin{table}[t!]
\caption{Ablation studies on different refinement on CMA.}
\setlength{\tabcolsep}{4pt}
\vspace{-0.1in}
\resizebox{\linewidth}{!}{
\begin{tabular}{c|ccccccc|c}
\toprule
\multirow{2}{*}{Method} & \multicolumn{7}{c|}{Testing Modality}                                                                                        & \multirow{2}{*}{Mean} \\ \cmidrule{2-8}
                        & F              & E              & L              & FE             & FL             & EL             & FEL            &                       \\ \midrule
MMD                     & 69.18          & 20.04 & 45.63          & 58.16          & 66.81          & 49.07          & 66.73          & 53.66                 \\
MSE                     & 62.93          & 18.27          & 42.17          & 53.77          & 61.22          & 51.62          & 64.15          & 50.59                 \\
InfoNCE                 & \textbf{70.56}          & \textbf{20.92}          & 42.27          & \textbf{63.48}          & 65.29          & 54.79          & 69.42          & 55.24                 \\
MMD + MSE               & 66.70 & 20.60          & \textbf{46.89} & 61.14 & \textbf{68.22} & \textbf{54.76} & \textbf{69.46} & \textbf{55.39}        \\ \bottomrule
\end{tabular}
}
\label{tab:CMA loss}
\vspace{-0.1in}
\end{table}

\begin{table}[t!]
\caption{Average feature distance with/without CMA.}
\setlength{\tabcolsep}{30pt}
\resizebox{\linewidth}{!}{
\begin{tabular}{ccc}
\toprule
Methods  & Modality Distance & Class Distance \\ \midrule
w/o. CMA & 0.34 & 5.23    \\
w. CMA   & \textbf{1.03} & \textbf{0.49}    \\ \bottomrule
\end{tabular}
}
\label{tab:distance}
\vspace{-0.2in}
\end{table}



\noindent \textbf{Ablations on the refinement in CMA.} Table \ref{tab:CMA loss} presents an ablation study evaluating different refinement strategies used in Cross Modality Alignment (CMA). The methods compared include MMD (Maximum Mean Discrepancy), MSE (Mean Squared Error), InfoNCE (Contrastive Learning Loss), and their combinations. The goal is to determine the most effective loss function for aligning cross-modal features. MMD achieves 53.66\% mIoU, indicating that distribution-level alignment provides moderate benefits but may lack instance-specific fine-tuning. MSE performs the worst (50.59\% mIoU), suggesting that minimizing direct feature differences alone is insufficient for effective refinement. InfoNCE leads to strong performance (55.24\% mIoU), demonstrating that contrastive learning effectively enhances feature alignment by maximizing inter-class separation and intra-class consistency. Combining MMD + MSE achieves the best performance (55.39\% mIoU), indicating that a hybrid approach leveraging both distribution alignment (MMD) and instance-level consistency (MSE) is optimal. These results highlight that contrastive learning (InfoNCE) is a strong refinement method while combining multiple regularization strategies (MMD + MSE) further enhances feature consistency.

Table \ref{tab:distance} compares the average and presents a comparison of modality distance and class distance with and without CMA. Modality distance measures the separation between different modalities, while class distance quantifies the intra-class feature compactness. Without CMA, the modality distance is low (0.34), indicating that features from different modalities collapse into a shared space, reducing modality-specific distinctions. Meanwhile, the class distance is high (5.23), suggesting poor intra-class consistency and dispersed feature representations. With CMA, modality distance significantly increases to 1.03, demonstrating that CMA effectively preserves modality-specific characteristics and prevents feature collapse. Simultaneously, class distance decreases sharply to 0.49, indicating that CMA enhances intra-class consistency by aligning features more closely to their respective semantic centers. These results confirm that CMA successfully strengthens modality separation while improving category-wise feature compactness.

\begin{table}[t!]
\caption{Comparasion on RGB only modality.}
\setlength{\tabcolsep}{40pt}
\resizebox{\linewidth}{!}{
\begin{tabular}{c|c|c}
\toprule
\multirow{2}{*}{Method} & \multirow{2}{*}{Backbone} & \multirow{2}{*}{mIoU} \\
                        &                           &                      \\ \midrule
CMNeXt                  & Seg-B0                    & 43.37                \\
Ours                    & Seg-B0                 & \textbf{60.03}       \\ \bottomrule
\end{tabular}
}
\label{tab:rgb}
\end{table}

\noindent \textbf{Performance on RGB only modality.} Table \ref{tab:rgb} compares the performance of different methods trained exclusively on the RGB modality. Our method, using ResNet-34, achieves a mIoU of 60.03\%, outperforming CMNeXt (43.37\% mIoU) with Seg-B0 as the backbone. This demonstrates that our approach effectively utilizes RGB information, achieving superior segmentation accuracy even without additional modalities.

\begin{table}[t]
\caption{Ablation studies on backbone initialization.}
\setlength{\tabcolsep}{4pt}
\vspace{-0.1in}
\resizebox{\linewidth}{!}{
\begin{tabular}{c|ccccccc|c}
\toprule
\multirow{2}{*}{Method} & \multicolumn{7}{c|}{Testing Modality}                                                                                        & \multirow{2}{*}{Mean} \\ \cmidrule{2-8}
                        & F              & E              & L              & FE             & FL             & EL             & FEL            &                       \\ \midrule
Random    & 58.58           & \textbf{24.98}           & 9.36           & 48.75           & 54.65           & 31.27           & 57.11           & 40.67                  \\
ImageNet     & \textbf{66.70} & 20.60 & \textbf{46.89} & \textbf{61.14} & \textbf{68.22} & \textbf{54.76} & \textbf{69.46} & \textbf{55.39}        \\ \bottomrule
\end{tabular}
}
\label{tab:init}
\vspace{-0.2in}
\end{table}

\noindent \textbf{Ablation Study on Backbone Initialization.} To assess the impact of backbone initialization, we compare models trained with random initialization and ImageNet pretraining, as shown in Table \ref{tab:init}. The results indicate that ImageNet pretraining significantly improves performance across all modalities, leading to a mean score increase from 40.67\% to 55.39\% (+14.72\%). When using random initialization, the model achieves relatively low performance, particularly in categories such as L (9.36\%) and FE (48.75\%), indicating difficulties in learning effective representations from scratch. In contrast, ImageNet initialization leads to notable improvements across all modalities, with F (66.70\%), L (46.89\%), FE (61.14\%), and FL (68.22\%) achieving substantial gains. This demonstrates that pretraining on large-scale datasets facilitates better feature extraction and improves generalization.


\begin{figure*}[t]
\begin{center}
\includegraphics[width=0.9\linewidth]{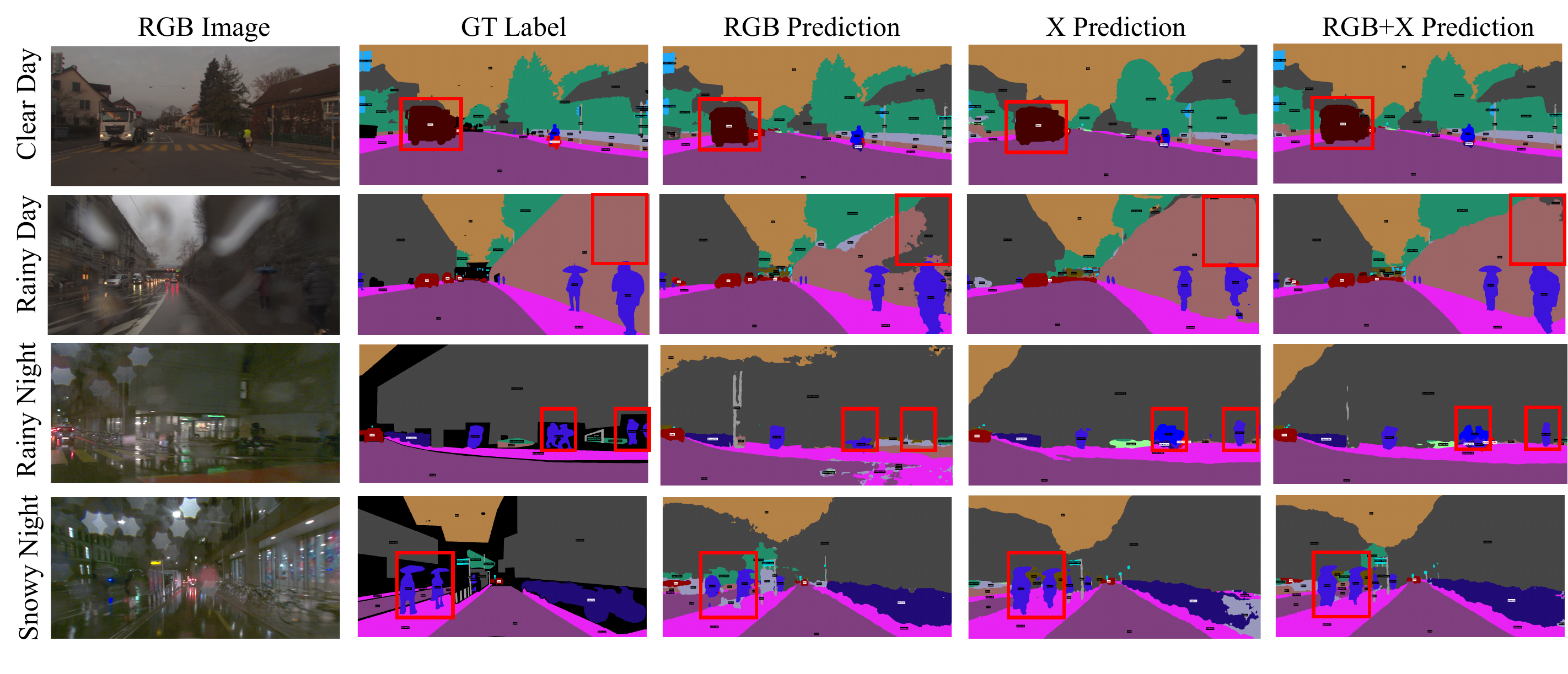}
\end{center}
\vspace{-0.3in}
\caption{Prediction of RGB and non-RGB modality under different conditions.}
\label{fig:both modality}
\vspace{-0.2in}
\end{figure*}

\begin{figure}[t]
\begin{center}
\includegraphics[width=1\linewidth]{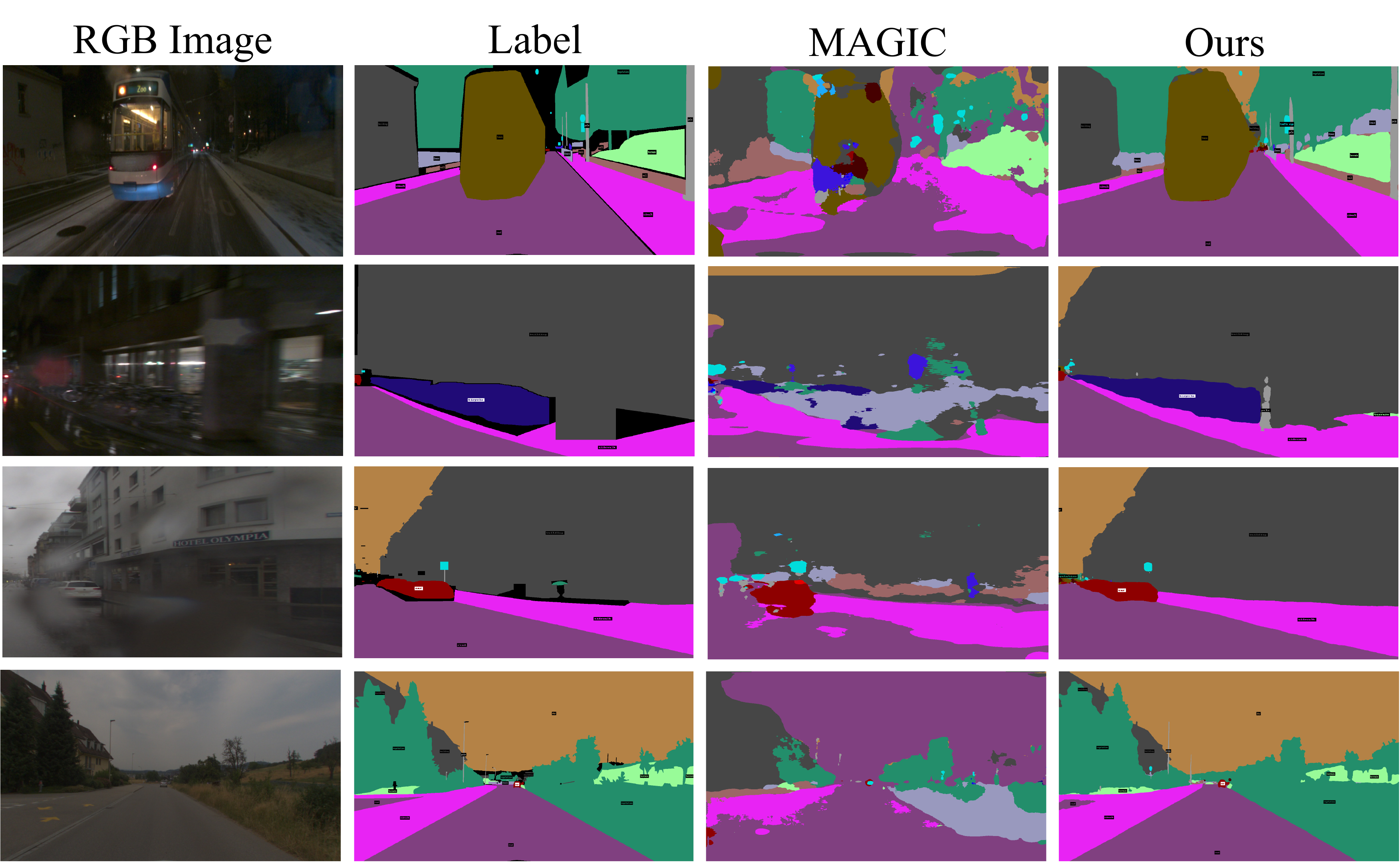}
\end{center}
\vspace{-0.1in}
\caption{Predictions visualization compared with MAGIC \cite{magic}.}
\label{fig:vis}
\vspace{-0.25in}
\end{figure}
\subsection{Qualitative Analysis}

\noindent \textbf{Contribution of RGB and X modality. } To evaluate our method's ability to maximize RGB and non-RGB modality's contribution, we visualize predictions under extreme conditions for RGB (Fig. \ref{fig:both modality}). The results highlight the strengths and limitations of different modalities: RGB and X modalities each specialize in their respective strengths, enabling more accurate segmentation; for example, depth improves boundary delineation, while event cameras better capture dynamic objects.

\noindent \textbf{Visualization of our method. }Fig. \ref{fig:vis} presents a qualitative comparison of semantic segmentation results between our method and MAGIC \cite{magic} across diverse driving scenarios, including low-light and adverse weather conditions. Our method demonstrates improved segmentation accuracy, with more precise results with the ground truth.

\section{Conclusion}
In this work, we propose \textbf{BiXFormer}, a novel robust framework for multi-modal semantic segmentation that effectively leverages the strength of different modalities. Unlike conventional methods that rely on feature fusion or knowledge distillation, BiXFormer preserves modality-specific advantages by reformulating the task as mask-level classification. Our proposed \textbf{Unified Modality Matching (UMM)} ensures optimal label matching through a two-step matching process, while \textbf{Cross Modality Alignment (CMA)} further refines modality-specific features by aligning strong and weak queries across modalities. By avoiding feature fusion and explicitly addressing missing modality scenarios, BiXFormer achieves significant improvements in segmentation accuracy. Extensive experiments on both synthetic and real-world benchmarks demonstrate the effectiveness of our approach, achieving substantial mIoU gains of \textbf{+2.75\%} and \textbf{+22.74\%} over prior methods. 

\section*{Acknowledgment}
Support for this work was given by the Toyota Motor Corporation (TMC) and JSPS KAKENHI Grant Number 23K28164 and JST CREST Grant Number JPMJCR22D1. However, note that this paper solely reflects the opinions and conclusions of its authors and not TMC or any other Toyota entity. Computations are done on the supercomputer “Flow” at the Information Technology Center, Nagoya University.

\bibliographystyle{IEEEtran}
\bibliography{ref}
\vspace{-0.5in}

\end{document}